\documentclass[conference]{IEEEtran}

\usepackage{color}
\usepackage{multirow,array,graphicx}
\usepackage{afterpage}
\usepackage{subfigure}
\usepackage{amsmath}
\usepackage{url}
\usepackage[linesnumbered,ruled,vlined,boxed]{algorithm2e}
\usepackage{multirow}
\usepackage{placeins}

\newcolumntype{P}[1]{>{\centering\arraybackslash}p{#1}}
\newcolumntype{M}[1]{>{\centering\arraybackslash}m{#1}}

\definecolor{darkgreen}{rgb}{0.0, 0.2, 0.13}
\definecolor{darkolivegreen}{rgb}{0.33, 0.42, 0.18}

\newcommand{\eg}{\textit{e.g.,}\ }
\newcommand{\ie}{\textit{i.e.,}\ }

\newcommand{\bioneat}{ \textsc{BioNeat}}

\renewcommand\IEEEkeywordsname{Keywords}
\title{Exploring Self-Assembling Behaviors in a Swarm of Bio-micro-robots using Surrogate-Assisted MAP-Elites}

\author{\IEEEauthorblockN{Leo Cazenille} \IEEEauthorblockA{Department of Information Sciences\\Ochanomizu University\\Tokyo, Japan\\cazenille.leo@ocha.ac.jp} \and \IEEEauthorblockN{Nicolas Bredeche} \IEEEauthorblockA{Sorbonne Universit\'{e}, CNRS\\Institut des Syst\`{e}mes Intelligents\\et de Robotique (ISIR)\\F-75005 Paris, France
\\nicolas.bredeche@sorbonne-universite.fr} \and \IEEEauthorblockN{Nathanael Aubert-Kato} \IEEEauthorblockA{Department of Information Sciences\\Ochanomizu University\\Tokyo, Japan\\naubertkato@is.ocha.ac.jp}}
\begin{document}

\maketitle

%-----------------------------------------------------------------------

\begin{abstract}
Swarms of molecular robots are a promising approach to create specific shapes at the microscopic scale through self-assembly. 
However, controlling their behavior is a challenging problem as it involves complex non-linear dynamics and high experimental variability. Hand-crafting a molecular controller will often be time-consuming and give sub-optimal results.  Optimization methods, like the \bioneat\ algorithm, were previously employed to partially overcome these difficulties, but they still had
to cope with deceptive high-dimensional search spaces and computationally expensive simulations.

Here, we describe a novel approach to solve this problem by using MAP-Elites, an algorithm that searches for both high-performing and diverse solutions. We then apply it to a molecular robotic framework we recently introduced that allows sensing, signaling and self-assembly at the micro-scale and show that MAP-Elites outperforms previous approaches.
Additionally, we propose a surrogate model of micro-robots physics and chemical reaction dynamics to reduce the computational costs of simulation. We show that the resulting methodology is capable of optimizing controllers with similar accuracy as when using only a full-fledged realistic model, with half the computational budget.
\end{abstract}

\begin{IEEEkeywords}%
Bio-micro-robots, swarm robotics, molecular programming, evolutionary robotics, surrogate models, quality-diversity algorithms, MAP-Elites
\end{IEEEkeywords}

\IEEEpeerreviewmaketitle

\section{Introduction} \label{sec:intro}
Swarms are impressive examples of collective behavior leading to the emergence of new properties~\cite{camazine2003self,sumpter2006principles,couzin2009collective}.
One of the most visible characteristics of swarms is their display of complex, dynamic structures~\cite{goodenough2017birds,anderson2002self,copeland2009bacterial}. Here, we are specifically interested in the creation of structure through cooperative assembly. Such behavior has been investigated both in the natural world~\cite{anderson2002self,copeland2009bacterial} as well as in robotic swarms~\cite{gross2006,tuci2006,ogrady2009,rubenstein2014}.

In this paper, we are interested in the automated programming of a swarm of micro-robots ($\gg1000$).
The micro-robots are microscopic agarose beads functionalized with bio-molecules (single strand DNA of 12-24 base pairs)~\cite{gines2017microscopic}. The beads forming the body of the robots are small enough to move through Brownian motion. The DNA functionalization allows them to produce (a) chemical signals that may impact the behavior of nearby beads and (b) an anchoring signal that will attach them to their neighbors. Clusters created from aggregated beads move slower or even stop, based on their size.
That property allows us to control where the beads should go, which we use to create self-assembled structures.  Thanks to that scale and the low price and availability of the molecular components, we previously designed and validated \emph{in vitro} a system of a million micro-robots that self-assemble~\cite{aubert2017evolutionary,zadorin2017synthesis}.

We  previously introduced the\bioneat\ algorithm~\cite{aubert2017evolutionary}, inspired from state-of-the-art \textsc{Neat} algorithm~\cite{stanley2002evolving}. \bioneat\ aims at producing chemical reaction networks (CRN) which represent bio-molecules to be attached to beads or left floating in the 2-dimensional substrate. Given a target for self-assembly (\eg in Fig.~\ref{fig:setup-targets}), the objective is that micro-robots assemble to one another in the target area. 

Though initial results are promising, this previous work raised several important issues:
\begin{itemize}
    \item \textbf{target complexity}: optimization was limited to simple targets, with one single rectangular region for self-assembly.
    \item \textbf{computational cost}: simulating chemical reactions in a realistic fashion is extremely expensive. Several days of computation were necessary for running the optimization algorithm.
\end{itemize}

In this paper, we address both issues. Firstly, we introduce the use of MAP-Elites~\cite{mouret2015illuminating}, an illumination algorithm that favors exploration over pure optimization, thus reducing the risk of premature convergence during the search. We also implement a new mechanism, termed \emph{topological initialization} that helps MAP-Elites to bootstrap exploration when confronted with search spaces with large neutral regions.

Secondly, we introduce a surrogate model~\cite{queipo2005surrogate} to approximate the (costly) pseudo-realistic simulation of real chemical reactions. We also introduce transferability~\cite{Koos2013transferability} in combination with the proposed surrogate model, \ie bootstrapping optimization using the surrogate model and then reverting to the costly pseudo-realistic simulator.

In the following, we first describe the Methods, both in terms of experimental setup, objective function and optimization algorithms. Then, we present the results obtained using a target function that combines several regions of interest. Finally, we discuss the remaining challenges of the approach and conclude.

\section{Methods} \label{sec:methods}
\subsection{Molecular robotic swarm}
\begin{figure}
\centering
\includegraphics[width = \linewidth]{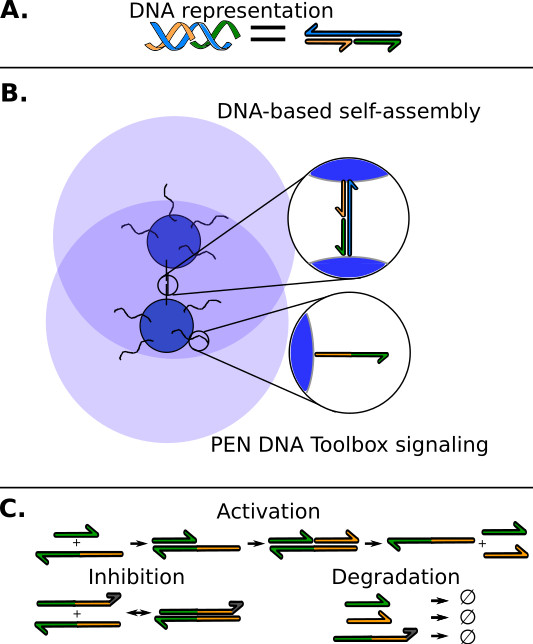}
\caption{\textbf{A.} Model of DNA: the 3D structures is abstracted, keeping only the orientation of the molecule. Here, two strands (orange and green) are double-stranded to a longer blue strand. \textbf{B.} Model of the robots. Two robots are shown here, both producing and emitting a DNA signal (blue diffusion area). The DNA strands grafted on their surface are used both for aggregation and signalling. \textbf{C.} Signalling mechanism. DNA strands diffusing in the environment interact with those grafted on the body of the robots, either producing other DNA strands (activation) or preventing production (inhibition). Signal strands are slowly degraded over time by an enzyme present in the environment. Strands grafted on the surface of the robot are chemically protected against this enzymatic activity and remain stable over time.}
\label{beads}
\end{figure}

We consider a swarm comprised of molecular robots made from spherical beads grafted with specific strands of DNA (Fig.~\ref{beads}). Those strands are of two types: (a) templates used to capture signal DNA from the environment, process it and produce signal back; (b) anchoring points, which are used for aggregation when the anchoring signal is present.

Processing and production of signals are based on the PEN DNA toolbox~\cite{montagne2011programming} (Fig.~\ref{beads}, c.): a signal strand can attach to a compatible template and either triggers the production of a different DNA strand (activation) or prevents activation from other strands (inhibition). The activation mechanism works as follows: once a compatible signal strand is double-stranded to a template, it gets extended into a fully complementary strand by a first enzyme called polymerase. The extend strand is designed to contain the recognition site of a second enzyme, called nickase, which splits it in half. That activity recovers the original signal strand (green in Fig.~\ref{beads}, c.) as well as produces a new signal strand (orange in Fig.~\ref{beads}, c.). That structure is unstable by design and both strands eventually fall off, allowing them to interact with other templates in the system. The inhibition mechanism relies on a signal strand attaching to a target template in a way that does not trigger the activity of either enzymes: a 3'-end mismatch prevents polymerase extension, while missing bases on the 5'-end of the strand prevents recognition by the nickase. At the same time, the inhibiting strand physically prevents activation signals to interact with the template, effectively inactivating it. As with activation signal, this structure is designed to be unstable, allowing the template to be eventually freed.

Signal strands are spread in the environment through chemical diffusion and are degraded over time by a third enzyme called exonuclease. Templates are chemically protected against that activity, thus remaining stable.

Additionally to templates, anchor strands are grafted on the beads. A specific signal, called anchoring signal, can link together anchor strands from two separate beads to link them together (Fig.~\ref{beads}, b.). Past experimental results showed that this mechanism is strong enough to produce stable aggregation \cite{aubert2017evolutionary}.

Robots move through Brownian motion: individual robots move much faster than aggregated clusters, which in turn may become completely motionless when they get large enough.

\begin{figure}
\begin{center}
\includegraphics[width=0.99\linewidth]{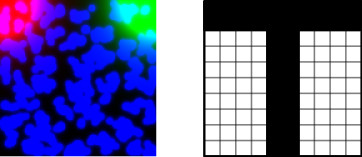}
\end{center}
\caption{\textbf{Left}: Simulation of an experiment where a swarm of micro-beads (in blue) interact with their environment and self-aggregate. Two gradient sources, in the top-left and top-right positions, are continuously diffused isotropically in the environment. They are shown respectively in red and green. \textbf{Right}: Target for self-assembly, with the shape of the letter T. Beads should aggregate in the black area.}
\label{fig:setup-targets}
\end{figure}

%%% %%% %%% %%% %%% %%% %%% %%% %%% %%% %%% 
%%% %%% %%% %%% %%% %%% %%% %%% %%% %%% %%% 
%%% %%% %%% %%% %%% %%% %%% %%% %%% %%% %%% 
%%% %%% %%% %%% %%% %%% %%% %%% %%% %%% %%% 
%%% %%% %%% %%% %%% %%% %%% %%% %%% %%% %%% 

\subsection{Simulation models} % a renommer "Evalution" (plus général)
We use two distinct models to simulate our molecular systems: a ``full'' model that explicitly computes beads positions and aggregation, and a ``surrogate'' model that considers a simplified scenario where beads are uniformly spread in the environment. 

In both cases, the production, diffusion, and degradation of signal species are modeled through reaction-diffusion. For a given signal $S$, we have:
$${\partial [S] \over \partial t} (t,x,y) = R_S(t,x,y) + D_S \cdot \Delta[S](t,x,y)$$

where $[S]$ is the concentration of $S$, $R_S$ is the contribution from reactions (\eg production or degradation, see Figure \ref{beads},C. for a graphical representation and \cite{aubert2014computer} for an explicit formula), $D_S$ is the diffusion coefficient of $S$ and $\Delta$ is the Laplacian operator. That approach generates a set of Partial Differential Equations (PDEs) that we solve numerically.

In the full model, beads are modelled as disks moving in the environment through Brownian motion. We implement it by adding Gaussian noise to their position over time, with mean $0$ and variance $\sigma^2 = 2D$ where:
$$D = { k_B T \over 6\pi d \eta_{\text{water}}}$$

with $k_B$ the Boltzmann constant, $T$ the temperature, $d$ the
bead diameter and $\eta_{\text{water}}$ the viscosity of water. To integrate with the reaction-diffusion part of the system, that Gaussian noise is scaled by the length of a time step from the PDE solver.

While the simulation is 2-dimensional, we consider that the beads are moving in a 3-dimensional environment, which allows us to ignore collisions. In presence of the anchoring signal, beads have a probability to aggregate. That probability is computed by implementing a Gillespie-like step: considering the current concentration of signal, we predict when the next aggregation event will occur~\cite{gillespie1976general}. If that event happens before the next time step of the PDE solver, aggregation is considered successful. The reverse reaction, a bead separating from an aggregate, is handled the same way. Since reactions probability distributions are memoryless, that approach does not change the overall behavior of the system. Once aggregated, we consider a cluster of $N$ moves in a similar fashion to a single bead with a radius $N$ times larger.

Note that keeping track of the behavior of each bead is computationally intensive. To minimize costs, we use far fewer beads in simulations than there would be in wetlab experiments. To still cover enough of the environment, a larger radius is used both to compute signal production and aggregation range. However, that radius would make the beads extremely slow; to keep the system dynamic, we consider that the beads move as if they had a smaller radius. Those two radii are indicated as \textit{bead size (aggregation)} and \textit{bead size (Brownian motion)} respectively in Table~\ref{tab:paramsSim}.

The local concentration of each signal-producing species is directly proportional to the number of beads at a given point of space. That is, we sum the concentrations of DNA molecules grafted on all the beads that are present. Note that, due to the non-linearity of the enzymatic reactions involved, a linear increase in signal-producing species does not necessarily mean a linear increase in the production of signal species.

In the surrogate model, we use the same reaction-diffusion approach, however the concentrations of signal-producing strands are considered spatially uniform and are equal to the respective concentrations of species present on the surface of a single bead. In practice, this model is equivalent to having motionless beads uniformly spread out in the environment or having the signal-producing species grafted on the surface of the environment rather than on beads. Note that such reaction-diffusion system can be experimentally implemented as well~\cite{padirac2013predator}.

\begin{figure}[h]
\begin{center}
\includegraphics[width=0.75\linewidth]{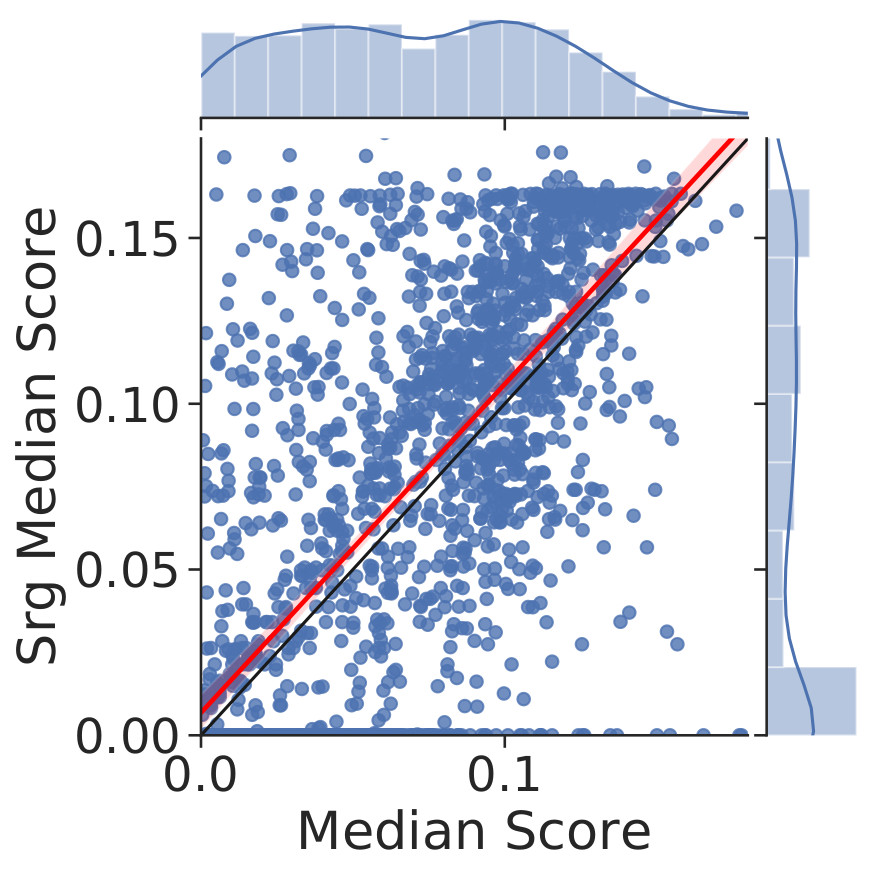}
\caption{Comparison of fitness evaluations results between the full model and the surrogate model across 5000 randomly generated individuals. A linear regression of the data is shown in red. The curve in black corresponds to the reference $x=y$. The surrogate model is shown to be an approximate of the basic model ($RMSE=0.0550$), with a weak (noisy) conservation of score ordering.}
\label{fig:approxComparison2}
\end{center}
\end{figure}

One of the main difference between these two models is that the full model is stochastic, due to the random motion of beads and probabilistic aggregations, while the surrogate model is deterministic. This characteristic makes the surrogate model much cheaper to compute as multiple repeats are not required. At the same time, not having to simulate the Brownian motion and aggregation of beads gives us an additional cut in computational cost. On the negative side, the surrogate model may fail to capture the lack of robustness of a given system. It is also incapable of taking into account possible designs relying on the increase in local concentrations when beads aggregate. However, we have found that, for the task presented in this paper, performances achieved with the simulated model or with the full model and with the surrogate model are correlated (Fig.~\ref{fig:approxComparison2} and Results section).

%%% %%% %%% %%% %%% %%% %%% %%% %%% %%% %%% 
%%% %%% %%% %%% %%% %%% %%% %%% %%% %%% %%% 
%%% %%% %%% %%% %%% %%% %%% %%% %%% %%% %%% 
%%% %%% %%% %%% %%% %%% %%% %%% %%% %%% %%% 
%%% %%% %%% %%% %%% %%% %%% %%% %%% %%% %%% 

\subsection{Target Objective} \label{sec:fitness}
We extend our work from~\cite{aubert2017evolutionary,cazenille2019swarm} with a more challenging problem, described in this paragraph. We previously devised a methodology to optimize CRN controllers driving a swarm of micro-beads to self-assemble into simple shapes composed of only one straight line. Here, we extends this approach further by focusing on a more complex type of pattern, composed of two lines. This requires that micro-robots to consider several chemical signals in order to perform a collection of attraction and repulsion behaviors to navigate to the target area.

We define a target for self-assembly corresponding to the shape of the letter \textbf{T} (Fig.~\ref{fig:setup-targets} right). Our objective is for the micro-beads to self-assemble into the target area, shown in black, starting from their initial random positions. Two fixed gradient sources are arranged in the top-left and top-right positions in the arena (Fig.~\ref{fig:setup-targets} left). They each emit a respective type of signal strand that diffuse throughout the environment. They provide the micro-beads information about their localization, potentially inducing self-assembly. As the gradients diffuse isotropically, self-assembling into straight patterns can be difficult.

We quantify the performance of a simulation by using the method described in~\cite{aubert2017evolutionary,cazenille2019swarm}.
The experimental arena is discretized into a $N \times N$ matrix of cells, with $N=160$. This allows to compute the following \textit{match-nomatch} score:

\begin{equation*}
f = \underbrace{\sum_{(\widetilde{x},\widetilde{y}) \in T} r * B(\widetilde{x},\widetilde{y})}_{\text{reward term}} 
- \underbrace{\sum_{(\widetilde{x},\widetilde{y}) \notin T}  p * e^{a * d(\widetilde{x},\widetilde{y})}* B(\widetilde{x},\widetilde{y})}_{\text{penalization term}}
\end{equation*}

with $T$ the set of $(\widetilde{x},\widetilde{y})$~cells in the target area, $r$ a reward parameter, $B(\widetilde{x},\widetilde{y})$ a Bool function of the presence of beads at position~$(\widetilde{x},\widetilde{y})$, $d$ the distance of a cell towards the closest position in $T$ and $a$ a scaling parameter.
Individual are rewarded according to the number of cells within $T$ and penalized for cells outside of $T$, with penalization increasing with distance to the target area.
Table~\ref{tab:paramsSim} lists the parameters used for simulation and fitness computation.

\begin{figure}[h]
\begin{center}
\includegraphics[width=0.35\textwidth]{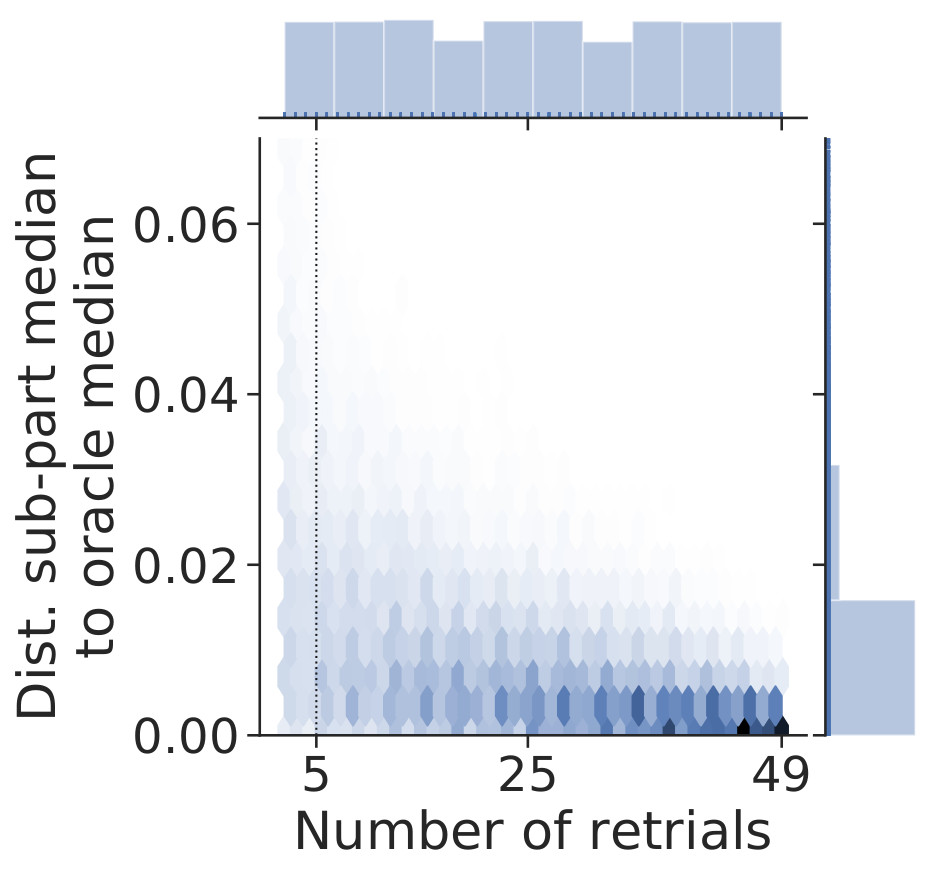}
\caption{Variability of evaluation scores on $5000$ randomly generated CRN individuals. Each individual is reevaluated $50$ times (\textit{oracle}). For each individual, we generate $1000$ random subsets, of size $2$ to $49$ from the \textit{oracle} sets of $50$ reevaluations. Medians of all subsets are compared to the respective medians of the \textit{oracle}. We postulate that 5 reevaluations per individuals is a good trade-off between precision and computational cost, and define our fitness metric as the median score across $N=5$ reevaluations of the same individual.}
\label{fig:statsRetrials}
\end{center}
\end{figure}

We define two fitness functions catering respectively to the full and surrogate models.
The surrogate model is deterministic, so the fitness is equal to the \textit{match-nomatch} score over one simulation.
The full model is stochastic. To alleviate simulation stochasticity, we assess the performance of an individual CRN by reevaluating $5$ simulations, computing their respective \textit{match-nomatch} scores, and defining the resulting fitness value as the median of these scores.
The selection of the number of reevaluations corresponds to a trade-off between precision and computational costs. Our choice of $5$ reevaluations was motivated by the results presented in Fig.~\ref{fig:statsRetrials} assessing fitness variability depending on the number of reevaluations.

Individuals are deemed valid only if their respective topology matches the requirements of Table~\ref{tab:params} in term of number of signal species and number of activation and inhibition templates.

\begin{table}[b]
\caption{Simulations and fitness parameters.}
\small
\begin{center}
\begin{tabular}{|p{3.5cm}|p{3.5cm}|}\hline
 Simulation Parameter & Value \\
 \hline
  Arena size & $1mm \times 1mm$ \\
  Beads & $500$ \\
  Bead size (aggregation) & $50 \mu m$ \\
  Bead size (Brownian motion) & $5 \mu m$ \\
  Temperature & $43^\circ C$ \\
  Grid size & $160 \times 160$ \\
  Time discretization & $0.1$ min per step \\
  Simulation duration & $4000$ steps (\ie $400$ min) \\
  Target width & $0.20 mm$ (20\% of arena width) \\
 \hline
\end{tabular}

\begin{tabular}{|p{3.5cm}|p{3.5cm}|}\hline
 Fitness Parameter & Value \\
 \hline
  $r$ (reward) & $1.0$ \\
  $p$ (penalty) & $0.2$ \\
  $a$ (scaling) & $0.1$ \\
 \hline
\end{tabular}
\end{center}
\label{tab:paramsSim}
\end{table}

%%% %%% %%% %%% %%% %%% %%% %%% %%% %%% %%% 
%%% %%% %%% %%% %%% %%% %%% %%% %%% %%% %%% 
%%% %%% %%% %%% %%% %%% %%% %%% %%% %%% %%% 
%%% %%% %%% %%% %%% %%% %%% %%% %%% %%% %%% 
%%% %%% %%% %%% %%% %%% %%% %%% %%% %%% %%% 

\subsection{Optimization} \label{sec:optimizers}
We use a two-step experimental protocol for optimization. As a first step, we perform both structural and parametric optimization to obtain chemical reaction networks (CRNs) through two possible methods: \bioneat\ and Map-Elites. Structural optimization modifies which templates are added to the beads, thus changing the set of reactions. Parametric optimization focuses on optimizing continuous values, namely DNA sequences stabilities and template concentrations. After convergence or exhaustion of the evaluation budget, we refine the designs obtained by performing a parametric-only optimization using CMA-ES\cite{Hansen2006}, a state-of-the-art evolutionary optimization method for continuous values.

\bioneat\ is an evolutionary algorithm we first introduced in~\cite{dinh2014effective}. It takes inspiration from the famous state-of-the-art \textsc{Neat} algorithm~\cite{stanley2002evolving}, which was originally designed to optimize artificial neural networks. \bioneat\ uses specific variation operators to navigate the search space of chemical reaction networks. It is also capable to protect innovation, that is, to explore several regions of the search space simultaneously, balancing between novelty of a particular design and the quality of solutions. We previously showed that \bioneat\ provides efficient solutions for targets comprised of a single regions (horizontal or vertical lines, see~\cite{aubert2017evolutionary} for details). The main limit of \bioneat\ is that while it can protect innovation for some time, there is not guarantee that it is capable of escaping completely the curse of premature convergence due to the pressure for optimizing towards (possibly only temporary) better solutions. 

\begin{table}[b]
\caption{Parameters of \bioneat\ and MAP-Elites.}
\begin{center}
\begin{tabular}{|p{5.7cm}|p{1.4cm}|}\hline
 Optimization Parameter & Value \\
 \hline
  Evaluation Budget & $3000$\\
  Number of retrials per individuals & $5$\\
  Maximal number of activation signals & $6$\\
  Maximal number of activation templates & $7$\\
  Maximal number of inhibition templates & $7$\\
 \hline
\end{tabular}

\begin{tabular}{|p{5.7cm}|p{1.4cm}|}\hline
 \bioneat\ Parameter & Value \\
 \hline
  Target number of \bioneat\ species & $20$ \\
  Population Size & $50$ \\
  Number of templates & $1 - 13$\\
 \hline
\end{tabular}

\begin{tabular}{|p{5.7cm}|p{1.4cm}|}\hline
 MAP-Elites Parameter & Value \\
 \hline
  Number of bins & $7$\\
  Batch size & $50$\\
  Number of elites per grid bin & $1$\\
  Number of templates & $7 - 13$\\
 \hline
\end{tabular}

\begin{tabular}{|p{5.7cm}|p{1.4cm}|}\hline
 Mutation operator & Probability \\
 \hline
  Parameter mutation & $0.80$\\
  Add template strand & $0.05$\\
  Remove template strand & $0.05$\\
  Add signal species & $0.05$\\
  Add inhibition species & $0.05$\\
 \hline
\end{tabular}
\end{center}
\label{tab:params}
\end{table}

\begin{algorithm}[h]
\DontPrintSemicolon
\SetAlgoLined
\SetAlgoNoEnd
\SetInd{0.1em}{1em}

\SetKwInOut{Input}{input}
\Input{$T_{min}, T_{max}$: min and max nr of templates}
\Input{$pmut_{min}$, $pmut_{max}$: min and max nr of parameter mutations}

$ind_{base} \leftarrow$ "3 signals (2 gradients + 1 anchoring signal), one autocatalitic activation template applied to the anchoring signal."\;

\While{True}{
    \tcc{Initialize random topology}
    ind $\leftarrow ind_{base}$\;
    $T_{target} \leftarrow $ randint($T_{min}$, $T_{max}$)\;
    \While{True}{
        \If{ind.nb\_templates $= T_{target}$}{
            break\;
        } \ElseIf{ind.nb\_templates $> T_{target}$}{
            disable\_template(ind)\;
        } \Else{
            apply\_topological\_mutations(ind)\;
        }
    }
    
    \tcc{Apply parameter mutations}
    $pmut_{target} \leftarrow $ randint($pmut_{min}$, $pmut_{max}$)\;
    \For{\_ in 1..$pmut_{target}$}{
        apply\_param\_mutation(ind)\;
    }

    \tcc{Verify if ind is valid}
    \If{ind.valid}{
        break\;
    }
}
return ind\;
\caption{Topological initialization of a CRN}
\label{alg:ti}
\end{algorithm}

\bioneat\ searches through CRN topologies iteratively. This behavior is inherited from \textsc{Neat} that postulated that iterated small changes in topologies would often only result in a mild effect on the fitness values. It may not be the case with CRN~\cite{yahiro2018reservoir}, where small changes in topology can have severe effects in fitness values, a particular trait of deceptive and hard-explore problems.
This may explain why \bioneat\ is prone to premature convergence during optimization, as it does not possess a way to explore a totally unexplored niche in the space of topologies. While this aspect is partially mitigated with \bioneat\ mechanism of speciation, which allows the parallel optimization of several topological niches, it does not enforce the discovery of totally novel niches. Worst, as new species are only created through atomic mutations (\ie change only a small part of the topology), the surviving species may contains individuals with very similar topologies, possibly in the same topological niche.

In order to favor exploration over pure optimization, we rely on MAP-Elites~\cite{mouret2015illuminating}, a Quality-Diversity algorithm~\cite{pugh2016quality}, that decomposes the search space into regions based on feature descriptions. It considers how does a candidate solution \textit{looks like} in the phenotypic space instead of considering how it is coded in the genotypic space. This method is particularly suited to cope with multi-modal, deceptive, hard exploration problems where traditional optimization algorithms would be prone to premature convergence, as in evolutionary robotics~\cite{mouret2015illuminating,cully2015robots,duarte2018evolution}.

MAP-Elites iteratively regroups the explored solutions in a grid of elites. This results in a collection of high-performing individuals across a number of features selected by the user, corresponding to the axes of the grid.
Here, we only consider a single feature corresponding to the total number of templates in the topology of an individual. 
CRN with smaller number of templates are easier to test experimentally but lose expressivity. Conversely, large-sized CRN can describe more complex behaviors, which may be necessary for the beads to successfully self-aggregate into the target shape. As such, a trade-off in term of topology complexity has to be considered, possibly after the optimization process. This substantiates methodologies that concurrently search for topologies of differing sizes.

MAP-Elites is equipped with the same set of mutation operators as \bioneat, and also retains its capability to optimize iteratively the topologies.
We introduce a novel methodology to bootstrap MAP-Elites exploration by initializing a collection of individuals with random topologies. This approach, described in Algorithm~\ref{alg:ti}, allows MAP-Elites to consider a large number of differing topologies. It makes use of the \bioneat\ mutation operators to generate individuals of varying topologies across a range of number of templates. In one optimization run, $10\%$ of the individuals are initialized with a random topology (\ie $300$ individuals for an evaluation budget of $3000$).
This contrasts with the \bioneat\ approach, where only small iterative changes in topologies are possible from mutations, and where individuals with totally new topologies are not initialized.

Table~\ref{tab:params} lists the chosen parameters for the MAP-Elites algorithm. We use our own implementation of \bioneat\ and MAP-Elites. \bioneat\ is open source and available from~\url{https://bitbucket.org/AubertKato/bioneat/}. MAP-Elites is coded in Python using the QDpy library~\cite{qdpy} and freely available as open source software at~\url{https://gitlab.com/leo.cazenille/qdpy}. All additional scripts used in this paper are available at~\url{https://bitbucket.org/leo-cazenille/daccad-qd}.

\afterpage{
\begin{table*}[h!]
\caption{Fitness scores of the best-performing individuals for all methods, across $100$ retrials. Those individuals are further optimized by CMA-ES and compared to their original with the Mann-Whitney U test. The $p$ value is shown in parenthesis.}
\begin{center}
\begin{tabular}{|c|M{2.0cm}|M{2.0cm}|M{2.0cm}|M{2.0cm}|M{2.0cm}|M{2.5cm}|}
 \hline
  Method & Full & Full+CMAES & srg & srg+CMAES & srg+trf & srg+tsf+CMAES \\
  \hline
  Expert & $0.52 \pm 0.05$ & $0.63 \pm 0.04$ ($p < 0.01$)& - & - & - & - \\
  \hline
  BN & $0.54 \pm 0.02$ & $0.60 \pm 0.02$ ($p < 0.01$) & $0.42 \pm 0.04$ & $0.62 \pm 0.03$ ($p < 0.01$)& $0.49 \pm 0.03$ & $0.51 \pm 0.04$ ($p < 0.01$) \\
  ME & $0.60 \pm 0.04$ & $0.63 \pm 0.04$ ($p < 0.01$) & $0.43 \pm 0.06$ & $0.55 \pm 0.03$ ($p < 0.01$) & $0.60 \pm 0.03$& $0.61 \pm 0.03$ ($p < 0.01$)\\
 \hline
\end{tabular}
\end{center}
\label{tab:scores}
\end{table*}

\begin{figure*}[h!]
\begin{center}
\includegraphics[width=1.00\textwidth]{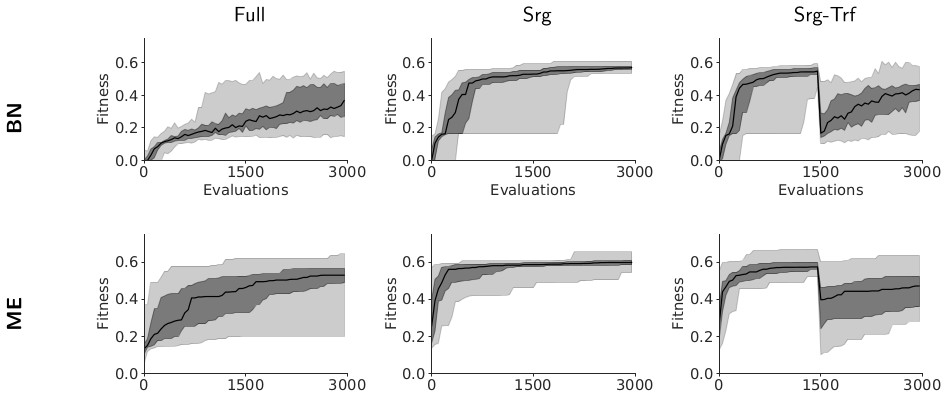}
\caption{Evolution of the median quality of the best-performing individuals for each target. Optimization methods are tested across $16$ different runs. A fitness of $1.0$ corresponds to the best performance. The darker shade represents the 25 to 75 percentiles. The lighter shade encompasses the minimal and maximal values. }
\label{fig:fitnessPerEvalsT}
\end{center}
\end{figure*}

\begin{figure*}[h]
\begin{center}

\begin{tabular}{c c c c c c c c}
& Expert & BN & BN+srg & BN+srg+tsf & ME & ME+srg & ME+srg+tsf \\

\parbox[t]{2mm}{\rotatebox[origin=c]{90}{}} &
\includegraphics[width=0.11\linewidth]{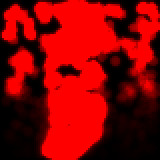} &
\includegraphics[width=0.11\linewidth]{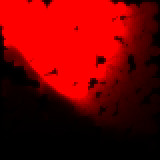} &
\includegraphics[width=0.11\linewidth]{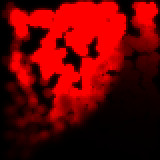} &
\includegraphics[width=0.11\linewidth]{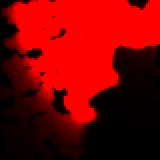} &
\includegraphics[width=0.11\linewidth]{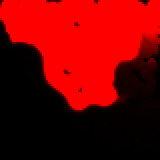} &
\includegraphics[width=0.11\linewidth]{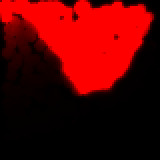} &
\includegraphics[width=0.11\linewidth]{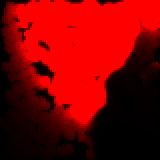} \\

\parbox[t]{2mm}{\vspace{-4em}\rotatebox[origin=c]{90}{CMA-ES}} &
\includegraphics[width=0.11\linewidth]{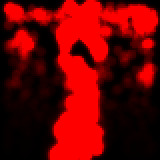} &
 \includegraphics[width=0.11\linewidth]{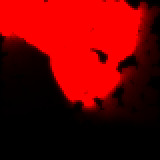} &%\includegraphics[width=0.11\linewidth]{figs/optimized_T_BioNEAT2-redComp--_VALID_.jpg} &
\includegraphics[width=0.11\linewidth]{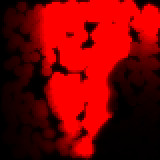} &
\includegraphics[width=0.11\linewidth]{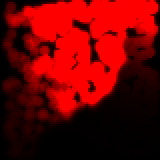} &
\includegraphics[width=0.11\linewidth]{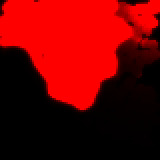} &
 \includegraphics[width=0.11\linewidth]{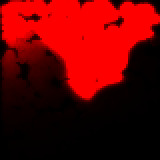} &%\includegraphics[width=0.11\linewidth]{figs/v2_optimized_approx_METI.jpg} &
\includegraphics[width=0.11\linewidth]{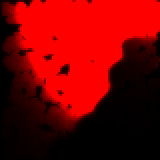} \\
\end{tabular}

\caption{Examples of final states obtained by \textbf{Full} simulations of the best-performing individuals optimized by each method, before and after calibration by CMA-ES.}
\label{fig:bestSolutions}
\end{center}
\end{figure*}

\begin{table*}[h!]
\caption{Mann-Whitney U test for comparing the distribution of the best individuals found by the different algorithms. The table shows which algorithm has the better mean between two given pairs and the $p$-value calculated by the test.}
%\footnotesize
\tiny
\begin{center}
\begin{tabular}{|p{1.5cm}|p{2.5cm}|p{2.5cm}|p{2.5cm}|p{2.5cm}|p{2.5cm}|}
\hline
 & BN+srg & BN+srg+tsf & ME & ME+srg & ME+srg+tsf  \\ 
 \hline
 BN & BN ($p = 7\times10^{-6}$)& BN+srg+tsf ($p = 0.12$) & ME ($p = 8\times10^{-4}$)& BN ($p = 10^{-5}$) & ME+srg+tsf ($p = 5\times10^{-4}$)\\
 BN+srg & - & BN+srg+tsf ($p = 9\times10^{-7}$)& ME ($p = 10^{-8}$)& ME+srg ($p = 0.19$)& ME+srg+tsf ($p = 6\times10^{-8}$)\\
 BN+srg+tsf & - & - & ME ($p = 2\times10^{-4}$)& BN+srg+tsf ($p = 2\times10^{-5}$)& ME+srg+tsf ($p = 9\times10^{-3}$)\\
 ME & - & - & - & ME ($p = 8\times10^{-8}$)& ME ($p = 0.10$)\\
 ME+srg & - & - & - & - & ME+srg+tsf ($p = 10^{-6}$)\\
 \hline
\end{tabular}
\label{tab:mannwhitney}
\end{center}
\end{table*}
\clearpage
}

\section{Results} \label{sec:results}
The\bioneat~and MAP-Elites algorithms (shorten as BN and ME for brevity) are used to optimize CRNs on the \textbf{T} target in three settings. 
The \textbf{full} and \textbf{srg} settings involve, respectively, the full and the surrogate simulations on a budget of $3000$ evaluations. The \textbf{srg+tsf} setting tests the transferability of the surrogate model: individuals are first optimized using the surrogate model during $1500$ evaluations. Then, all individuals of the last iteration (last population for \bioneat, final grid of MAP-Elites) are reevaluated and optimized using the full model over $1500$ evaluations.

All results are compared to a CRN designed by an expert, corresponding to the \textbf{Expert} setting.

The best-performing individuals of each method are reevaluated using the full model, with scores presented in Table~\ref{tab:scores}. For each method, CRN parameters of the best reevaluated individuals are optimized by CMA-ES~\cite{Hansen2006} over $2000$ evaluations, as in~\cite{aubert2017evolutionary}: the topology of these individuals are fixed, but templates concentration and activation signals stability parameters are optimized.

\textbf{Full} runs are computed in $15$ hours per run, \textbf{srg} runs take $90$ minutes per run and \textbf{srg+tsf} runs take $9$ hours per run.

For each setting, the evolution of fitness values during 16 optimization runs are presented in Fig.~\ref{fig:fitnessPerEvalsT}.

Figure~\ref{fig:bestSolutions} shows the final state of a full simulation of the best-performing solutions for each setting, before and after optimization by CMA-ES. Only the anchoring signal is represented, corresponding to self-assembled clusters of beads.

To compare the performance of the different settings, we collected the best individuals from each run (16 individuals per setting). For any two given settings, we used a Mann-Whitney U test on the distribution of fitness of their respective best individuals (Table~\ref{tab:mannwhitney}). The algorithms can be ordered in terms of mean performance: ME~$\simeq$ ME+srg+tsf~$\gg$ BN+srg+tsf ~$\simeq$ BN ~$\gg$ ME+srg~$\simeq$ BN+srg, where $\gg$ indicates a statistically significant difference ($p < 0.05$). The remaining cases, shown by a $\simeq$ have $p = 0.11$, $p = 0.12$ and $p = 0.19$ respectively.

In order to evaluate the impact of further parametric optimization refinement, we picked the best performing individual for each method, and compare performance \textit{before} and \textit{after} an extra parametric optimization step (\ie the topology is fixed) using the CMA-ES optimization algorithm. All individuals were reevaluated 100 times with the full model and the distribution of their evaluations were compared with a Mann-Whitney U test). 

Ordering the best individuals from each method \textit{before} parametric optimization gives the following: ME~$\simeq$ ME+srg+tsf~$\gg$
BN~$\gg$ Expert~$\gg$ BN+srg+tsf~$\gg$ ME+srg~$\gg$ BN+srg. 

\textit{After} parametric optimization, the order of the best individuals becomes:  Expert+CMAES~$\simeq$ ME+CMAES~$\gg$ BN+srg+CMAES~$\gg$ ME+srg+tsf+CMAES~$\gg$
BN+CMAES~$\gg$ ME+srg+CMAES~$\gg$
BN+srg+tsf+CMAES.

Finally, we also measured the impact of the optimization by CMA-ES by comparing the distribution of evaluations of a given individual and that of its optimized version. We used once again 100 retrials with a Mann-Whitney U test. The $p$-values were added to Table~\ref{tab:scores}. All optimized individuals were significantly better than their original ($p < 0.05$) except for the best individual of the ME with Full model setting ($p = 0.08$).

\begin{figure}
 Best individual before optimization with CMA-ES: ME
 
 \begin{minipage}{0.58\linewidth}
 \centering
 \includegraphics[width = \linewidth]{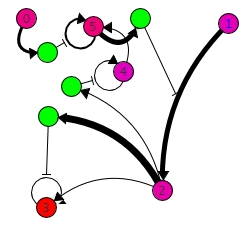}
 \end{minipage}\hfill
 \begin{minipage}{0.40\linewidth}
 \hfill\includegraphics[width = 0.8\linewidth]{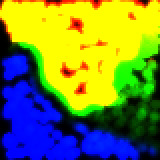}
 
 \hfill\includegraphics[width = 0.8\linewidth]{figs/v4_T-ME-TI-redComp--_VALID.jpg}
\end{minipage}

Best individual after optimization with CMA-ES: Expert

 \begin{minipage}{0.58\linewidth}
 \centering
 \includegraphics[width = \linewidth]{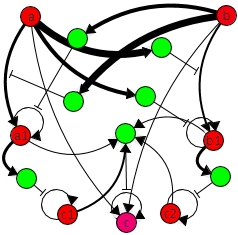}
 \end{minipage}\hfill
 \begin{minipage}{0.40\linewidth}
 \hfill\includegraphics[width = 0.8\linewidth]{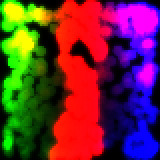}
 
 \hfill\includegraphics[width = 0.8\linewidth]{figs/optimized_HumanT4-redComp--_VALID_.jpg}
 \end{minipage}
 
\caption{Best-performing solutions before (top) and after (bottom) parameter tuning done by CMA-ES. Left: chemical reaction networks, vertices are signal strands (red: activating, green: inhibiting), edges are templates, inhibition is shown by a bar-headed arrow; top-right: chemical concentrations; bottom-right: production of anchoring signals.}
\label{fig:expertVSME}
\end{figure}

In accordance with the specifications of Table~\ref{tab:params}, optimized individuals are noticeably smaller than the expert-designed CRN (19 templates, Fig.~\ref{fig:expertVSME}): 13 for BN, 12 for BN+srg and BN+srg+tsf, 11 for ME, 10 for ME+srg and ME+srg+tsf. Fewer templates is actually beneficial with respect to potential \emph{in vitro} experimentation, as the models used loose accuracy when the number of chemical interactions grows.

\section{Discussion and conclusion}
We presented an approach to optimize CRNs driving the behavior of a swarm of bio-micro-robots to self-assemble into a \textbf{T}-shaped target.
Our approach improves upon the methodology in~\cite{aubert2017evolutionary}, based on\bioneat. We then described a method to optimize CRNs using the MAP-Elites~\cite{mouret2015illuminating}. It is a Quality-Diversity algorithm~\cite{pugh2016quality}, allowing us to explore solutions that are both high-performing and diverse. It is particularly suitable to cope with the deceptive nature of our target problem. We also introduced the notion of topological initializations to bootstrap MAP-Elites.
We showed that MAP-Elites outperforms\bioneat~both in term of performance and convergence speed. While the best-performing individuals optimized by\bioneat~still require an additional optimization stage using CMA-ES to provide competitive results, MAP-Elites does not necessitate this procedure.

We presented a way to reduce the computational costs of simulations through the use of a surrogate model of the robot swarm dynamics. Our optimization approach required far less computational power, while still retaining sufficient accuracy. We described how individuals optimized through the surrogate model can be transferred into optimization processes implementing the full model. We showed that such results are competitive with results obtained using only the full model, nearly dividing by two the optimization effort (60\%).

We limited the optimization runs to small CRN ($\leq$ 13 templates) to ease potential experimental validations. Indeed, smaller CRNs are easier to test experimentally, but lose expressivity. We showed that all methods are capable of optimizing small CRN while still being competitive with results obtained by the expert-designed CRN, that involves a far larger number of templates (19). Additionally, MAP-Elites allows for the concurrent optimization of CRN of differing size (number of templates), resulting in several CRNs that could be selected after the optimization process for experimentation, depending on the choices and requirements of the experimenter.

Our approach with MAP-Elites could be improved by testing additional features rather than only the number of templates, including topological features (\eg number of vertices, properties of the graph) or behavioral features (\eg occupation of various zones of the arena, speed of reaching a stable state).
The interplay between surrogate and full model could be improved further by finding automatically when to perform a transfer. The optimization process could also switch several times between surrogate and full model evaluations, either periodically or under some predetermined conditions.
Finally, our approach could be applied to more complex target shapes, possibly requiring the implementation of mechanisms to optimized CRN iteratively, or by automatically decomposing the target shapes into sub-shapes.

\section*{Acknowledgements}
\small
This work was supported by JSPS KAKENHI Grant Number JP17K00399 and by Grant-in-Aid for JSPS Fellows JP19F19722. This work was also supported by the Agence Nationale pour la Recherche under Grant ANR-18-CE33-0006 (MSR project).

%\bibliographystyle{plain}
%\bibliography{biblio} 
\FloatBarrier

\end{document}